\documentclass[runningheads]{llncs}
\usepackage{graphicx}
\usepackage{comment}
\usepackage{amsmath,amssymb} % define this before the line numbering.
\usepackage{color}
\usepackage{epsfig}
\usepackage{enumerate}
\usepackage{multirow}
\usepackage[FIGTOPCAP]{subfigure}
\usepackage{algorithm}
\usepackage{algorithmic}
\usepackage{varwidth}

\usepackage{cite}
\usepackage{bm}
\def \bbE{{\mathbb{E}}}
\def \bbR{{\mathbb{R}}}

\def \cF{{\mathcal{F}}}
\def \cL{{\mathcal{L}}}
\def \cN{{\mathcal{N}}}
\def \cT{{\mathcal{T}}}
\def \cX{{\mathcal{X}}}
\def \cY{{\mathcal{Y}}}
\def \cZ{{\mathcal{Z}}}

\def \bmt{{\bm{t}}}
\def \bmx{{\bm{x}}}
\def \bmz{{\bm{z}}} 

\def \bmmu{{\boldsymbol\mu}} 

\usepackage[font=small,labelfont=bf, belowskip=-3pt]{caption}
\begin{document}
% \linenumbers
\pagestyle{headings}
\mainmatter

\title{Leveraging Seen and Unseen Semantic Relationships for Generative Zero-Shot Learning} % Replace with your title

\begin{comment}
\titlerunning{ECCV-20 submission ID \ECCVSubNumber} 
\authorrunning{ECCV-20 submission ID \ECCVSubNumber} 
\author{Anonymous ECCV submission}
\institute{Paper ID \ECCVSubNumber}
\end{comment}
%******************

%\begin{comment}
\titlerunning{Leveraging Seen and Unseen Semantic Relationships for G-ZSL}
\author{Maunil R Vyas  \orcidID{0000-0003-3906-1540} \and 
Hemanth Venkateswara  \orcidID{0000-0002-3832-0881} \and
Sethuraman Panchanathan  \orcidID{0000-0002-8769-6340} 
}
\authorrunning{Vyas et al.}
\institute{Arizona State University, Tempe AZ 85281, USA \\
\email{\{mrvyas, hemanthv, panch\}@asu.edu}}
%\end{comment}
%******************
\maketitle

\begin{abstract}
Zero-shot learning (ZSL) addresses the unseen class recognition problem by leveraging semantic information to transfer knowledge from seen classes to unseen classes. 
Generative models synthesize the unseen visual features and convert ZSL into a classical supervised learning problem. 
These generative models are trained using the seen classes and are expected to implicitly transfer the knowledge from seen to unseen classes. 
However, their performance is stymied by overfitting, which leads to substandard performance on Generalized Zero-Shot learning (GZSL). To address this concern, we propose the novel LsrGAN, a generative model that Leverages the Semantic Relationship between seen and unseen categories and explicitly performs knowledge transfer by incorporating a novel Semantic Regularized Loss (SR-Loss). The SR-loss guides the LsrGAN to generate visual features that mirror the semantic relationships between seen and unseen classes. Experiments on seven benchmark datasets, including the challenging Wikipedia text-based CUB and NABirds splits, and Attribute-based AWA, CUB, and SUN, demonstrates the superiority of the LsrGAN compared to previous state-of-the-art approaches under both ZSL and GZSL. Code is available at \emph{\url{https://github.com/Maunil/LsrGAN}}

\end{abstract}

\section{Introduction} 
Consider the following discussion between a kindergarten teacher and her student.\\ 
\emph{ 
\textbf{Teacher}: Today we will learn about a new animal that roams the grasslands of Africa. It is called the Zebra. \\
\textbf{Student}: What does a Zebra look like ?\\
\textbf{Teacher}: It looks like a short white horse but has black stripes like a tiger.} \\\\
That description is nearly enough for the student to recognize a zebra the next time she sees it. 
The student is able to take the verbal (textual) description and relate it to the visual understanding of a horse and a tiger and generate a zebra in her mind. 
In this paper, we propose a zero-shot learning model that transfers knowledge from text to the visual domain to learn and recognize previously unseen image categories. 

Collecting and curating large labeled datasets for training deep neural networks is both labor-intensive and nearly impossible for many of the classification tasks, especially for the fine-grained categories in specific domains. 
Hence, it is desirable to create models that can mitigate these difficulties and learn not to rely on large labeled training sets. 
Inspired by the human ability to recognize object categories solely based on class descriptions and previous visual knowledge, the research community has extensively pursued the area of ``Zero-shot learning''  (ZSL) ~\cite{LNH13, LEB08, RSS11, YA10, Xu17, Ding17}. 
Zero-shot learning aims to recognize objects that are not seen during the training process of the model. 
It leverages textual descriptions/attributes to transfer knowledge from seen to unseen classes. 

\begin{figure}[t]
\centering
\includegraphics[width=0.6\textwidth,height=3.8cm]{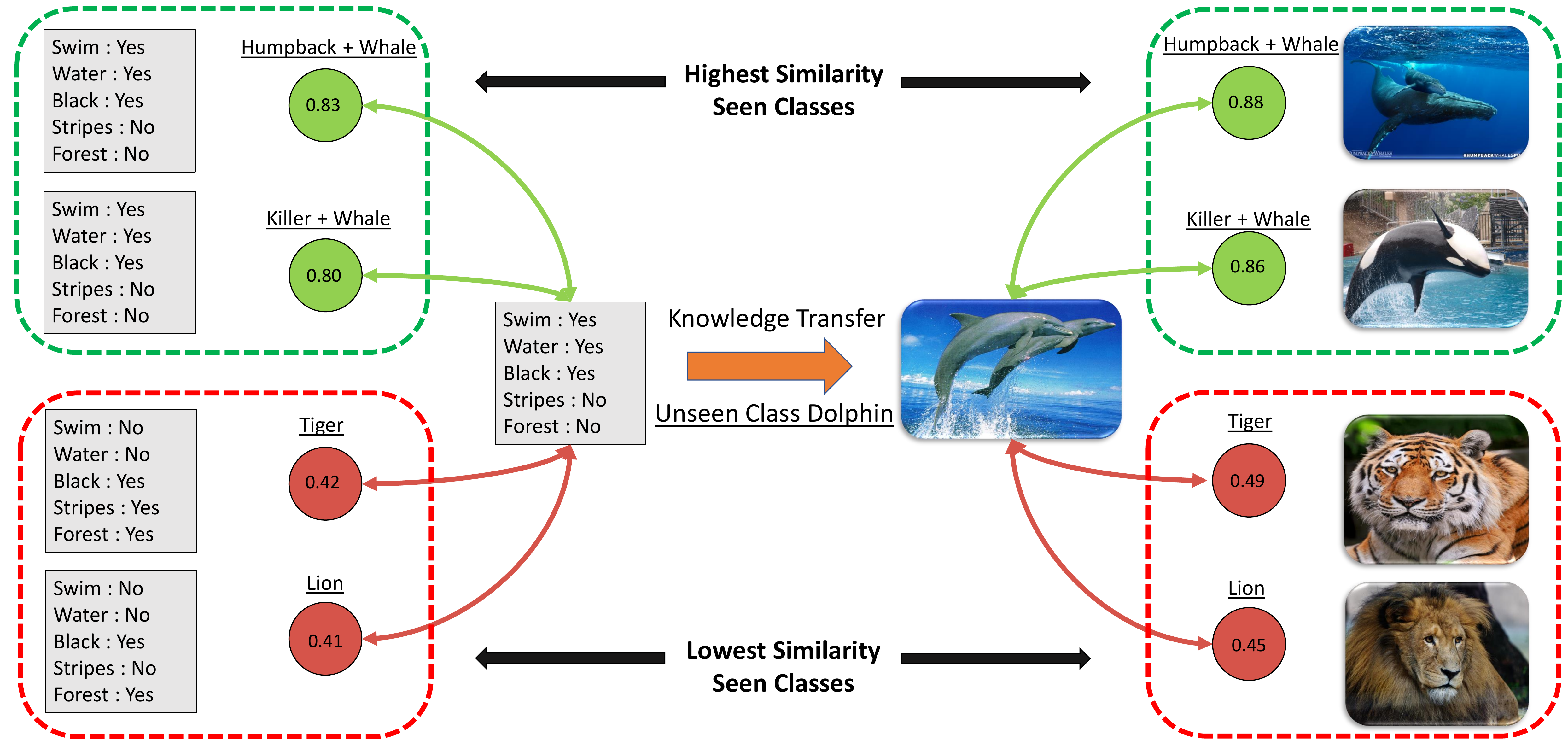}
\caption{Driving motivation behind leveraging the semantic relationship between seen and unseen classes to infer the visual characteristics of unseen classes. Notice that though the feature representations are different, the class similarity values are almost the same. e.g. ``Dolphin" has almost identical similarity values in visual and semantic space with other seen classes. The similarity values are mentioned in the circles, and computed using the cosine distance. }
%\setlength\belowcaptionskip{-0.20\baselineskip}
%\vspace{-15pt}
\label{fig:example1}
\end{figure}

Generative models are the most popular approach to solve zero-shot learning. 
Despite the recent progress, generative models for zero-shot learning still have some key limitations. 
%These models are trained only on the seen classes as the visual features for the unseen classes are not available . 
These models show a large quality gap between the synthesized and the actual unseen image features \cite{F-WGAN, VAE1, CycleGAN1, LGAN, Noisy_Text2}.
As a result, the performance of generalized zero-shot learning (GZSL) suffers, since many synthesized features of unseen classes are classified as seen classes. 
The second major concern behind the current generative model based approaches is the assumption that the semantic features are available in the desired form for a class category, e.g., clean attributes. 
However, in reality, they are hard to get. 
Getting clean semantic features requires a domain expert to annotate the attributes manually. 
Moreover, collecting a large number of attributes for all the class categories is again labor-intensive and expensive. 
Considering this, our generative model learns to transfer semantic knowledge from both noisy text descriptions (like Wikipedia articles) as well as semantic attributes for zero-shot learning and generalized zero-shot learning.

\indent In this paper we propose a novel generative model called the LsrGAN. 
The LsrGAN leverages semantic relationships between seen and unseen categories and transfers the same to the generated image features. We implement this transfer through a unique semantic regularization framework called the Semantic Regularized Loss (SR-Loss). 
In Fig. \ref{fig:example1}, ``Dolphin", an unseen class, has a high semantic similarity with classes such as ``Killer whale'' and ``Humpback whale" from the seen class set. 
These two seen classes are the potential neighbors of the Dolphin in the visual space. 
Therefore, when we do not have the real visual features for the Dolphin class, we can use these neighbors to form indirect visual references for the Dolphin class.  This illustrates the intuition behind the proposed SR-loss. The LsrGAN also trains a classifier to guide the feature generation. The main contributions of our work are:
\begin{enumerate}
\item A generative model leveraging the semantic relationship (LsrGAN) between seen and unseen classes overcoming the overfitting concern towards seen classes. 
\item A novel semantic regularization framework that enables knowledge transfer across the visual and semantic domain. The framework can be easily integrated into any generalized zero-shot learning model. 
\item We have conducted extensive experiments on seven widely used standard benchmark datasets to demonstrate that our model outperforms the previous state-of-the-art approaches.  
\end{enumerate}

\section{Related work}
%\vspace{-5pt}
Earlier work on ZSL was focused on learning the mapping between visual and semantic space in order to compensate for the lack of visual representation of the unseen classes. These approaches are known as \textit{Embedding methods}. The initial work was focused on two-stage approaches where the attributes of an input image are estimated in the first stage, and the category label is determined in the second phase using highest attribute similarity. DAP and IAP \cite{LNH13} are examples of such an approach.  Later, the use of bi-linear compatibility function led to promising ZSL results. Among these, ALE \cite{akata2015label} and DEVISE \cite{frome2013devise} use ranking loss to learn the bilinear compatibility function. ESZSL \cite{romera2015embarrassingly} applies the square loss and explicitly regularizes the objective w.r.t. the Frobenius norm. Unlike standard embedding methods, \cite{zhang2017learning, changpinyo2017predicting} propose reverse mapping from the semantic to the visual space and perform nearest neighbor classification in the visual space. The hybrid models such as CONSE \cite{norouzi2013zero}, SSE \cite{zhang2015zero}, and SYNC \cite{changpinyo2016synthesized}, discuss the idea of embedding both visual and semantic features into another intermediate space. These methods perform well in the ZSL setting. However, in GZSL, they show a high bias towards seen classes.  \\
\indent Most of the mentioned embedding methods have a bias towards seen classes leading to substandard GZSL performance. Recently, \textit{Generative methods} \cite{verma2017simple, chen2018zero, F-WGAN, LGAN,Noisy_Text2, CycleGAN1, VAE1, guo2017synthesizing} have attempted to address this concern by synthesizing unseen class features, leading to state of the art ZSL and GZSL performance. 
%Recent generative approaches for ZSL are mainly based on the VAE \cite{kingma2013auto} and GAN \cite{GAN1, WGAN1} based architectures. 
Among these, \cite{F-WGAN, LGAN, Noisy_Text2, CycleGAN1, chen2018zero} are based on Generative Adversarial Networks (GAN), while \cite{VAE1, bucher2017generating} use Variational Autoencoders (VAE) for the ZSL. F-GAN \cite{F-WGAN} uses a Wasserstein GAN \cite{WGAN, WGAN1} to synthesize samples based on the class conditioned attribute. LisGAN \cite{LGAN} proposes a \emph{soul-sample} regularizer to guide the sample generation process to stay close to a generic example of the seen class. Inspired by the cycle consistency loss\cite{zhu2017unpaired}, Felix et al. \cite{CycleGAN1}, propose a generative model that maps the generated visual features back to the original semantic feature, thus addressing the unpaired training issue during generation. These generative methods use the annotated attribute as semantic information for the feature generation. However, in reality, such a desired form of the semantic feature representation is hard to obtain. 
Hence, \cite{elhoseiny2013write} suggests the use of Wikipedia descriptions for ZSL and GZSL. GAZSL \cite{Noisy_Text2} proposes a very first generative model that handles the Wikipedia description to generate features. \\
\indent Although generative methods have been quite successful in ZSL, the unseen feature generation is biased towards seen classes leading to poor generalization in GZSL. For better generalization, we have proposed a novel SR-loss to leverage the inter-class relationships between seen and unseen classes in GANs. 
The idea of utilizing inter-class relationships has been addressed before with Triplet loss-based approaches \cite{annadani2018preserving, cacheux2019modeling} and using contrastive networks \cite{jiang2019transferable}. 
In this work, we propose a related approach using generative models. To best of our knowledge, such an approach using GANs has not been investigated.

%\vspace{-16pt}
\begin{figure}[t]
\centering
\includegraphics[width=0.73\textwidth,height=3.22cm]{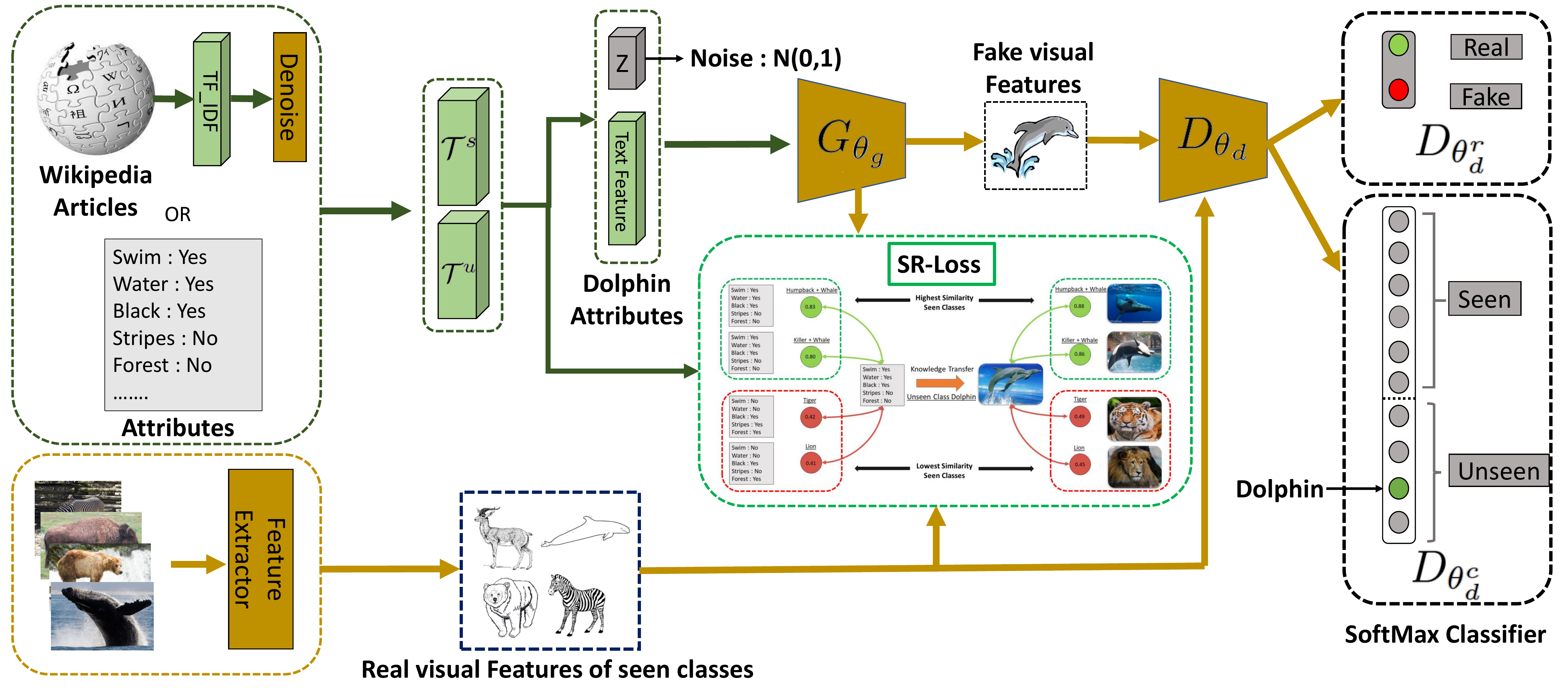}
\caption{Conceptual illustration of our LsrGAN model. The basis of our model is a conditional WGAN. The novel SR-Loss is introduced to help the $G_{\theta_{g}}$ to understand the semantic relationship between classes and guide it for applying the same during visual feature generation. The $G_{\theta_{g}}$ will use $\cT^s$ and $\cT^u$ to generate visual features. The $D_{\theta_{d}}$ has two branches used to perform real/fake game and classification. Notice that when we train the $G_{\theta_{g}}$ using $\cT^u$, only the classification branch remains active in $D_{\theta_{d}}$ as the unseen visual features are not available. }
%\vspace{-16pt}
\label{fig:example2}
\end{figure}

\section{Proposed Approach}
\subsection{Problem Settings}
%Jay shree satimaa
%Jay ho mira datar bapu 
%Jay shree swaminarayan bhagwan 
%Jay shree ram 
%Jay shree hanuman 

The zero-shot learning problem consists of seen (observed) and unseen (unobserved) categories of images and their corresponding text information. 
Images belonging to the seen categories are passed through a feature extractor (for e.g., ResNet-101) to yield features $\{\bmx_i^s\}_{i=1}^{n_s}$, where $\bmx \in \cX^s$. 
The corresponding labels for these features are $\{y_i^s\}_{i=1}^{n_s}$, where $y^s \in \cY^s = \{1,\ldots, C_s\}$ with $C_s$ seen categories. 
The image features for the unseen categories are denoted as $\{\bmx_i^u\}_{i=1}^{n_u}$, where $\bmx \in \cX^u$, and the space of all image features is $\cX := \cX^s \cup \cX^u$. 
As the name indicates, unseen categories are not observed, and the zero-shot learning model attempts to hallucinate these features with the rest of the information provided. 
Although we do not have the image features for the unseen categories, we are privy to the $C_u$ unseen categories, where the corresponding labels for the unseen image features would be $\{y_i^u\}_{i=1}^{n_u}$, with $y^u \in \cY^u = \{C_s+1,\ldots, C\}$, where $C = C_s + C_u$. 
From the text domain, we have the semantic features for all categories which are either binary attribute vectors or Term-Frequency-Inverse-Document-Frequency (TF-IDF) vectors. 
The category-wise semantic features are denoted as $\{\bmt_c^s\}_{c=1}^{C_s}$, for seen categories and $\{\bmt_c^u\}_{c=C_s+1}^C$ with $\bmt \in \cT := \cT^s \cup \cT^u$. 
The goal of zero-shot learning is to build a classifier $\cF_{zsl} : \cX^u  \to \cY^u$, mapping image features to unseen categories, and the goal of the more difficult problem of generalized zero-shot learning is to build a classifier $\cF_{gzsl} : \cX  \to \cY := \cY^s \cup \cY^u$, mapping image features to seen and unseen categories. 

\subsection{Adversarial Image Feature Generation} 
Using the image features of the seen categories and the semantic features of the seen and unseen categories, we propose a generative adversarial network to hallucinate the unseen image features for each of the unseen categories. 
We apply a conditional Wasserstein Generative Adversarial Network (WGAN) to generate image features for the unseen categories using semantic features as input \cite{WGAN}. The WGAN aligns the real and generated image feature distributions. 
In addition, we have a feature classifier that is trained to classify image features into $C$ categories of seen and unseen classes. 
The components of the WGAN are described in the following.\\ 

\noindent \textbf{Feature Generator}: The conditional generator in the WGAN has parameters $\theta_g$ and is represented as $G_{\theta_g} : \cZ \times \cT \to \cX$, where $\cZ$ is the space of random normal vectors $\cN(0,I)$ of $|\cZ|$ dimensions. 
%$|\cT|$ and $|\cX|$ are the dimensions of the semantic and image feature spaces, respectively. 
Since the TF-IDF features from the Wikipedia articles may contain repetitive and non-discriminating feature information, we apply a denoising transformation upon the TF-IDF vector using a fully-connected neural network layer as proposed by \cite{Noisy_Text2}. 
The WGAN takes as input a random noise vector $\bmz \in \cZ$ concatenated with the semantic feature vector $\bmt_c$ for a category $c$, and generates an image feature $\tilde{\bmx}_c \leftarrow G_{\theta_g}(\bmz, \bmt_c)$. 
The generator is trained to generate image features for both seen categories ($\tilde{\bmx}_c^s \leftarrow G_{\theta_g}(\bmz, \bmt^s_c)$) and unseen categories ($\tilde{\bmx}_c^u \leftarrow G_{\theta_g}(\bmz, \bmt^u_c)$). 
In order to generate image features that are structurally similar to the real image features, we implement visual pivot regularization, $\cL_{vp}$ that aligns the cluster centers of the real image features with the cluster centers of the generated image features for each of the $C_s$ categories \cite{Noisy_Text2}. 
This is implemented only for the seen categories where we have real image features. 
\begin{flalign}
\cL_{vp} = \min_{\theta_g} \frac{1}{C_s}\sum_{c=1}^{C_s}\big|\big| \bbE_{(\bmx,y=c) \sim (\cX^s, \cY^s)}[\bmx] - \bbE_{(\bmz,\bmt^s_c) \sim (\cZ, \cT^s)}[G_{\theta_g}(\bmz, \bmt_c^s)] \big|\big|.
\label{Eq:Gen_vp}
\end{flalign}

\noindent \textbf{Feature Discriminator}: We train the WGAN with an adversarial discriminator having two branches to perform the real/fake game and classification. The discriminator has parameters $\theta_{d}^r$ and $\theta_{d}^c$ for two branches respectively and is denoted as $D_{\theta_d}$. The real/fake branch of the discriminator learns a mapping $D_{\theta_d^r}: \cX \to \bbR $ using the generated and real image features to output $D_{\theta_d^r}(\tilde{\bmx}^s)$ and $D_{\theta_d^r}(\bmx^s)$ that are used to estimate the objective term $\cL_d$. 
The objective $\cL_d$ is maximized w.r.t. the discriminator parameters $\theta_{d}^r$ and minimized w.r.t. the generator parameters $\theta_g$. 
\begin{flalign}
\cL_d = \min_{\theta_g} \max_{\theta_d} \bbE_{\bmx \sim \cX^s}\big[D_{\theta_d}(\bmx) \big] - \bbE_{(\bmz, \bmt) \sim (\cZ, \cT^s)}\big[D_{\theta_d}(G_{\theta_g}(\bmz, \bmt))\big] - \lambda_{gp}\cL_{gp} 
\label{Eq:Disc_Dd}
\end{flalign}
where, the first two terms control the alignment of real image feature and the generated image feature distributions. 
The third term is the gradient penalty to enforce the Lipschitz constraint with $\mathcal{L}_{gp} = (||\nabla_{\bmx} D_{\theta_d}(\bmx)||_2 - 1)^2$ where, input $\bmx$ are real image features, generated image features and random samples on a straight line connecting real image features and generated image features \cite{WGAN1}. 
The parameter $\lambda_{gp}$ controls the importance of the Lipschitz constraint. 
The discriminator parameters are trained using only seen category image features since $\bmx^u$ are unavailable.\\ 

\noindent \textbf{Feature Classifier}: The category classifier has parameters $\theta_{d}^c$ and is denoted as $D_{\theta_d^c}$. 
It is a softmax cross-entropy classifier for the generated and real image features $\bmx^s, \tilde{\bmx}^s, \tilde{\bmx}^u$ and is trained to minimize the loss $\cL_{c}$. For ease of notation we represent the real/generated image features as $\bmx$ and corresponding labels $y$

\begin{flalign}
\cL_c = \min_{{\theta_g}, {\theta_d^c}} -\bbE_{(\bmx,y) \sim (\cX, \cY)}\bigg[\sum_{c=1}^C 1(y=c)\text{log}(D_{\theta_d^c}(\bmx))_c \bigg], 
\label{Eq:Disc_Dc}
\end{flalign}

where $(D_{\theta_d^c}(\bmx))_c$ is the $c$-th component of the $C$-dimension softmax output and $1(y=c)$ is the indicator function. While the discriminator performs a marginal alignment of real and generated features, the classifier performs category based conditional alignment.\\ 
\indent So far, in the proposed WGAN model the \emph{Generator} generates image features from seen and unseen categories. 
The \emph{Discriminator} aligns the image feature distributions and the \emph{Classifier} performs image classification. 
The LsrGAN model is illustrated in Fig. \ref{fig:example2}. 
In the following we introduce a novel regularization technique that transfers knowledge across the semantic and image feature spaces for the unseen and seen categories respectively. 

\subsection{Semantic Relationship Regularization} 
%The key component behind the generative methods for zero-shot learning is the generator $G_{\theta_g}$. Conventional zero-shot learning approaches only use the seen classes in the training process when generating image features \cite{F-WGAN,Noisy_Text2,LGAN}. This hinders the learning capability of the generator since there is no knowledge about the unseen classes during the training phase. Moreover, this also leads to overfitting the generator towards the seen classes leading to poor performance in generalized zero-shot learning. We aim to mitigate these issues by proposing a novel regularization procedure that will explicitly transfer knowledge of unseen classes from the semantic domain and guide the generator in generating seen and unseen image features.  We term this the ``Semantic Regularized loss (SR-Loss)''. 
% Modified with the above one!!
Conventional generative zero-shot learning approaches \cite{F-WGAN,Noisy_Text2,LGAN}, have poor generalization performance in GZSL since the generated visual features are biased towards the seen classes. 
We address this issue by proposing a novel regularization procedure that will explicitly transfer knowledge of unseen classes from the semantic domain to guide the model in generating distinct seen and unseen image features. 
We term this the ``Semantic Regularized Loss (SR-Loss)''.

We argue that the visual and semantic feature spaces share a common underlying latent space that generates the visual and semantic features. We propose to exploit this relationship by transferring knowledge from the semantic space to the visual space to generate image features. Knowing the inter-class relationships in the semantic space can help us impose the same relationship constraints among the generated visual features. This is the idea behind the SR-Loss in the WGAN where we transfer inter-class relationships from the semantic domain to the visual domain. Fig. 1 illustrates this concept.  The visual similarity between class $c_i$ and $c_j$ is represented as $\cX_{sim}(\bmmu_{c_i}, \bmmu_{c_j})$, where $\bmmu_c$ is the mean of the image features of class $c$. 
Note that for visual similarity we are considering the relationship between the class centers and not between individual image features. Likewise, the semantic similarity between class $c_i$ and $c_j$ is represented as $\cT_{sim}(\bmt_{c_i}, \bmt_{c_j})$. 
For semantic similarity, we do not have the mean, since we only have one semantic vector $\bmt_c$ for every category, although the proposed approach can be extended to include multiple semantic feature vectors. We impose the following semantic relationship constraint for the image features, 
\begin{flalign}
\cT_{sim}(\bmt_{c_i}, \bmt_{c_j}) - \epsilon_{ij} ~ \leq ~ \cX_{sim}(\bmmu_{c_i}, \bmmu_{c_j}) ~ \leq ~ \cT_{sim}(\bmt_{c_i}, \bmt_{c_j}) + \epsilon_{ij}, 
\label{Eq:Lsr1}
\end{flalign}
where, hyper-parameter $\epsilon_{ij} \geq 0$ is a soft margin enforcing the similarity between semantic and image features for classes $i$ and $j$. 
Large values of $\epsilon_{ij}$ allow more deviation between semantic similarities and visual similarities. We incorporate the constraint into the objective by applying the penalty method \cite{penalty},  
\begin{flalign}
p_{c_{ij}} & \big[|| \max (0,  \cX_{sim}(\bmmu_{c_i}, \bmmu_{c_j})  -  (\cT_{sim}(\bmt_{c_i}, \bmt_{c_j}) +  \epsilon_{ij})) ||^2 \notag\\
          & + || \max (0, (\cT_{sim}(\bmt_{c_i}, \bmt_{c_j}) -  \epsilon_{ij}) -  \cX_{sim}(\bmmu_{c_i}, \bmmu_{c_j})) ||^2\big], 
\label{Eq:Lsr2}
\end{flalign}
where, $p_{c_{ij}}$ is the penalty for violating the constraint. 
The penalty becomes zero when the constraints are satisfied and is non-zero otherwise. We use cosine distance to compute the similarities.

We intend to transfer semantic inter-class relationships to enhance the image feature representations that are output from the generator. 
Consider a seen class $c_i$. 
We estimate its semantic similarity $\cT_{sim}(\bmt_{c_i}, \bmt_{c_j})$ with all the other seen semantic features $\bmt_{c_j}$ where $j \in \{1, \ldots, C_s\} \land j \neq i$. 
Not all the similarities are important and for the ease of implementation we select the highest $n_c$ similarities that we wish to transfer. 
Let $I_{c_i}$ represent the set of $n_c$ seen categories with the highest semantic similarity with $c_i$. 
We apply Eq \ref{Eq:Lsr2} to train the generator to output image features that satisfy the semantic similarity constraints against the top $n_c$ similarities from the seen categories. 
For the seen image categories, the objective function is, 
\begin{flalign}
\cL_{sr}^s = \min_{\theta_g} \frac{1}{C_s}&\sum_{i=1}^{C_s}\sum_{j\in I_{c_i}}\big[|| \max (0,  \cX_{sim}(\bmmu^s_{c_j}, \tilde{\bmmu}^s_{c_i})  -  (\cT_{sim}(\bmt^s_{c_j}, \bmt^s_{c_i}) +  \epsilon)) ||^2 \notag\\
          & + || \max (0, (\cT_{sim}(\bmt^s_{c_j}, \bmt^s_{c_i}) -  \epsilon) -  \cX_{sim}(\bmmu^s_{c_j}, \tilde{\bmmu}^s_{c_i})) ||^2\big], 
\label{Eq:Lsr3}
\end{flalign}
where, penalty $p_{ij} = 1$, and $\bmmu^s_{c_j} := \bbE_{(\bmx,y=c_j)\sim (\cX^s,\cY^s)}[\bmx]$ is the mean of the image features for seen class $c_j$, and $\tilde{\bmmu}^s_{c_i} := \bbE_{\bmz \sim \cZ}[G_{\theta_g}(\bmz, \bmt^s_{c_i})]$ is the mean of the generated image features of seen class $c_i$. 
We use a constant value of $\epsilon$ as the soft margin to simplify the solution. 
Similarly, the objective function for the unseen categories is, 
\begin{flalign}
\cL_{sr}^u = \min_{\theta_g} \frac{1}{C_u}&\sum_{i=C_s+1}^{C}\sum_{j\in I_{c_i}}\big[|| \max (0,  \cX_{sim}(\bmmu^s_{c_j}, \tilde{\bmmu}^u_{c_i})  -  (\cT_{sim}(\bmt^s_{c_j}, \bmt^u_{c_i}) +  \epsilon_{ij})) ||^2 \notag\\
          & + || \max (0, (\cT_{sim}(\bmt^s_{c_j}, \bmt^u_{c_i}) -  \epsilon_{ij}) -  \cX_{sim}(\bmmu^s_{c_j}, \tilde{\bmmu}^u_{c_i})) ||^2\big], 
\label{Eq:Lsr4}
\end{flalign}
where, $\tilde{\bmmu}^u_{c_i} := \bbE_{\bmz \sim \cZ}[G_{\theta_g}(\bmz, \bmt^u_{c_i})]$ is the mean of the generated image features of unseen class $c_i$. The time complexity of the proposed SR-loss is $\mathcal{O}(B n_c |\cX| E)$ + $\mathcal{O}(C^2 |\cT| E)$ + $\mathcal{O}(C^2 \log{} n_c E)$ where $B$, $n_c$, $|\cX|$, $E$, $C$ and $|\cT|$ denote the batch size, neighbor size, number of visual features, epoch, number of total classes and the size of the semantic features respectively. This shows the computation cost is manageable.

%In terms of the time complexity, the proposed loss is linear with Epoch, Batch size, the dimension of the visual and semantic feature, and the number of neighbors. The cost of finding the cosine similarity among the classes is degree 2 polynomial with the class numbers. This showcases the computation cost is manageable. More details are mentioned in the appendix. 

\subsection{LsrGAN Objective Function} 
The LsrGAN leverages the semantic relationship between seen and unseen categories to generate robust image features for unseen categories using the objective function defined in Eq. \ref{Eq:Lsr3} and \ref{Eq:Lsr4}. The model generates robust seen image features conditioned by the regularizer in Eq. \ref{Eq:Gen_vp}. 
The LsrGAN trains a classifier over all the $C$ categories as outlined in Eq. \ref{Eq:Disc_Dc}. 
The LsrGAN is based on a WGAN model that aligns the image feature distributions using the objective function defined in Eq. \ref{Eq:Disc_Dd}. 
The overall objective function of the LsrGAN model is given by, 
\begin{flalign}
\lambda_c\cL_c + \cL_d + \lambda_{vp}\cL_{vp} + \lambda_{sr}(\cL^s_{sr} + \cL^u_{sr})
\label{Eq:LsrGAN}
\end{flalign}
where, $\lambda_c$, $\lambda_{vp}$ and $\lambda_{sr}$ are hyper parameters controlling the importance of each of the loss terms. 
Unlike standard zero-shot learning models that generate image features and then have to train a supervised classifier \cite{F-WGAN, Noisy_Text2, LGAN}, the LsrGAN model has an inbuilt classifier that can also be used for evaluating zero-shot learning and generalized zero-shot learning.  

\section{Experiments}
%Table 1 and Table 2 show the ZSL and GZSL results. We can see that the proposed LsrGAN outperformed the previous models.  

\subsection{Datasets}
\begin{table*}[t]
\centering
\caption{Dataset Information. For the attribute-based datasets, the (number) in seen classes denotes the number of  classes used for test in GZSL.} 
\label{tab:gzsl2}
\resizebox{\textwidth}{!}{
\small\begin{tabular}{|l|ccc|cccc|}
\hline
\multirow{4}{*}{} & \multicolumn{3}{c|}{attribute-based} & \multicolumn{4}{c|}{Wikipedia descriptions}\\
\hline
%& \multicolumn{2}{c|}{CUB} & \multicolumn{2}{c|}{NAB} & \multicolumn{2}{c|}{CUB} & \multicolumn{2}{c|}{NAB}  \\
 & AWA & CUB & SUN & CUB (Easy) & CUB (Hard) & NAB (Easy) & NAB (Hard) \\
\hline %\cline{2-13}
No. of Samples &  30,475 & 11,788 & 14,340 & 11,788 & 11,788 & 48,562 & 48,562 \\ 
No. of Features &  85 & 312 & 102 & 7551 &  7551 & 13217 & 13217 \\
No. of Seen classes &  40(13) & 150(50) & 645(65) & 150 & 160 & 323 & 323 \\
No. of Unseen classes &  10 & 50 & 72 & 50 & 40 & 81 & 81 \\
\hline
\end{tabular}
}
%\vspace{-12pt}
\end{table*}

\textbf{Attribute-based datasets :}
We conduct experiments on three widely used attribute-based datasets : \textit{Animal with Attributes} (AWA) \cite{LNH13}, \textit{Caltech-UCSD-Birds 200-2011} (CUB) \cite{CaltechUCSDBirdsDataset} and Scene UNderstanding (SUN) \cite{PH12}. AWA is a medium scale coarse-grained animal dataset having 50 animal classes with 85 attributes annotated. CUB is a fine-grained, medium-scale dataset having 200 bird classes annotated with 312 attributes. SUN is a medium scale dataset having 717 types of scenes with 102 annotated attributes. We followed the split mentioned in \cite{good_ugly} to have a fair comparison with existing approaches. 

\noindent \textbf{Wikipedia descriptions-based datasets:} 
In order to address a more challenging ZSL problem with Wikipedia descriptions as auxiliary information, we have used two common fine-grained datasets with textual descriptions: CUB and \textit{North America Birds} (NAB) \cite{NAB}. The NAB dataset is larger compared to CUB having 1011 classes in total. We have used two splits, suggested by \cite{Elhoseiny_2017_CVPR} in our experiments to have a fair comparison with other methods. The splits are termed as \textit{Super-Category-Shared} (SCS, Easy split)  and \textit{Super-Category-Exclusive} (SCE, Hard split). These splits represent the similarity between seen and unseen classes. The SCS-split has at least one seen class for every unseen class belonging to the same parent. For example,  ``Harris's Hawk" in the unseen set and ``Cooper's Hawk" in the seen set belong to the same parent category, ``Hawks." On the other hand, in the  SCE-split, the parent categories are disjoint for the seen and unseen classes. Therefore, SCS and SCE splits are considered as Easy and Hard splits. The details for each dataset and class splits are given in Table 1. 
\indent \subsection{Implementation Details and Performance Metrics}
The 2048-dimensional ResNet-101 \cite{ResNet-101} features are considered as a real visual feature for attribute-based datasets, and part-based features  (e.g., belly, leg, wing, etc.) from VPDE-net\cite{VPDE} are used for the Wikipedia based datasets, as suggested by \cite{Noisy_Text2, F-WGAN}. We have utilized the TF-IDF to extract the features from the Wikipedia descriptions. For a fair comparison, all of our experiment settings are kept the same as reported in \cite{good_ugly, Noisy_Text2, F-WGAN}. \\
\indent The base block of our model is GAN, which is implemented using a multi-layer perceptron. Specifically, the feature generator $G_{\theta_g}$ has one hidden unit having 4096 neurons and LeakyReLU as an activation function. For attribute-based datasets, we intend to get the top max-pooling units of ResNet - 101 (visual features). Hence, the output layer has ReLU activation in the feature generator. On the other hand, for the Wikipedia based datasets, we have used Tanh as an output activation for the feature generator since the VPDE-net feature varies from -1 to 1. $\cZ$ is sampled from the normal Gaussian distribution. To perform the denoising and dimensionality reduction from Wikipedia descriptions, we have employed a fully connected layer with a feature generator. Also, notice that the semantic similarity for the SR-Loss is computed using the denoiser's output in Wikipedia based datasets. We will discard this layer when dealing with attribute-based datasets. In our model, the discriminator $D_{\theta_{d}}$ has two branches. One is used to play the real/fake game, and the other performs the classification on the generated/real visual feature. The discriminator also has 4096 units in the hidden layer with ReLU as an activation. Since the cosine distance is less prone to the curse of dimensionality when the features are sparse (semantic features), we have considered it in the SR-loss.  \\
\indent To perform the zero-shot recognition we have used nearest neighbor prediction on datasets having Wikipedia descriptions, and the classifier attached to the discriminator for the attributes based recognition. Notice that the classifier is not re-trained for the recognition part. The Top-1 accuracy is used to assess the ZSL setting. Furthermore, to capture the more realistic scenario, we have examine the Generalized zero-shot recognition as well. As suggested by \cite{AUC_Score}, we report the area under the seen and unseen curve (AUC score) as GZSL performance metric for Wikipedia based datasets, and the  
harmonic mean of the seen and unseen Top-1 accuracies for attribute based dataset.  Notice that the choice of these measures and predictions models is to make a fair comparison with existing methods.

\begin{table*}[t]
\centering
\caption{ZSL and GZSL results on AWA, CUB, and SUN with attributes as semantic information. T1 indicates the Top-1 \% accuracy in the ZSL setting. On the other hand, ``U", ``S" and ``H" denotes the Top-1\% accuracy for the unseen, seen, and Harmonic mean (seen + unseen). } 
\label{tab:gzsl2}
\resizebox{0.9\textwidth}{!}{
\small\begin{tabular}{|l|ccc|ccc|ccc|ccc|}
\hline
 & \multicolumn{3}{c|}{zero-shot Learning} & \multicolumn{9}{c|}{Generalized zero-shot Learning}\\
\cline{2-13}
Methods & AWA & CUB & SUN & \multicolumn{3}{c|}{AwA} & \multicolumn{3}{c|}{CUB} & \multicolumn{3}{c|}{SUN}  \\
\cline{2-13}
%\hline
\cline{2-13}
 & T1 & T1 & T1 & U & S & H & U & S & H & U & S & H  \\
 \hline %\cline{2-13}
DAP~\cite{LNH13} & 44.1 & 40.0 & 39.9 & 0.0 & 88.7 & 0.0 & 1.7 & 67.9 & 3.3 & 4.2 & 25.2 & 7.2  \\
%\hline
CONSE~\cite{norouzi2013zero}  & 45.6 & 34.3 & 38.8 &  0.4 & 88.6 & 0.8 & 1.6 & 72.2 & 3.1 & 6.8 & 39.9 & 11.6 \\
%\hline
SSE~\cite{zhang2015zero} & 60.1 & 43.9 & 51.5 & 7.0 & 80.5 & 12.9 & 8.5 & 46.9 & 14.4 & 2.1 & 36.4 &4.0  \\     
%\hline
DeViSE~\cite{frome2013devise} & 54.2 & 50.0 & 56.5 & 13.4 & 68.7 & 22.4 & 23.8 & 53.0 & 32.8 &16.9 & 27.4 & 20.9  \\

SJE~\cite{akata2015evaluation} & 65.6 & 53.9 & 53.7 & 11.3 & 74.6 & 19.6 & 23.5 & 59.2 & 33.6 & 14.7 &30.5 & 19.8  \\
%\hline
ESZSL~\cite{romera2015embarrassingly} & 58.2 & 53.9 & 54.5  & 5.9 & 77.8 & 11.0 & 2.4 & 70.1 & 4.6 &11.0 &27.9 & 15.8  \\
%\hline
ALE~\cite{akata2015label} & 59.9 &54.9 & 58.1 & 14.0 & 81.8 & 23.9 & 4.6 & 73.7 & 8.7 &21.8 &33.1 &26.3  \\     
%\hline
SYNC~\cite{changpinyo2016synthesized} &54.0 &55.6 & 56.3 & 10.0 & {\bf 90.5} & 18.0 & 7.4 & 66.3 & 13.3 & 7.9 & \textbf{43.3} & 13.4    \\

SAE~\cite{kodirov2017semantic} &53.0 &33.3 &40.3 & 1.1 & 82.2 & 2.2 & 0.4 & {\bf 80.9} & 0.9 & 8.8 &18.0 & 11.8   \\
%\hline
DEM~\cite{zhang2017learning} & 68.4 &51.7 & 61.9 & 30.5 & 86.4 & 45.1 & 11.1 & 75.1 & 19.4 &20.5 &34.3 &25.6\\

PSR ~\cite{annadani2018preserving} & 63.8 & 56.0 & 61.4 & 20.7 & 73.8 & 32.3 &  24.6 & 54.3 & 33.9 &20.8 &37.2 &26.7\\

TCN ~\cite{jiang2019transferable} & 70.3 & 59.5 & 61.5 & 49.4 & 76.5 & 60.0 &  52.6 & 52.0 & 52.3 &31.2 &37.3 &34.0\\
\hline
GAZSL~\cite{Noisy_Text2} & 68.2 & 55.8 & 61.3 & 19.2 & 86.5 & 31.4 & 23.9 & 60.6 & 34.3 & 21.7 & 34.5 & 26.7  \\     
%\hline
F-GAN~\cite{F-WGAN} &68.2 &57.3 &60.8 & {\bf 57.9} & 61.4 & 59.6 & 43.7 & 57.7 & 49.7 & 42.6 & 36.6 & 39.4  \\
cycle-CLSWGAN ~\cite{CycleGAN1} &66.3 &58.4 &60.0 & 56.9 & 64.0 & 60.2 & 45.7 & 61.0 & 52.3 & \textbf{49.4} & 33.6 & 40.0 \\
LisGAN \cite{LGAN} & \textbf{70.6} & 58.8 & 61.7 & 52.6 & 76.3 &  62.3 & 46.5 & 57.9 & 51.6 & 42.9 &37.8 &40.2  \\
\hline
LsrGAN [ours] & 66.4 & \textbf{60.3} & \textbf{62.5} &  54.6 & 74.6 & \textbf{63.0} & \textbf{48.1} & 59.1 & \textbf{53.0} &  44.8 & 37.7 & \textbf{40.9} \\
\hline
\end{tabular}
}
%\vspace{-8pt}
\end{table*}

\begin{table*}[t]
\centering
\caption{ZSL and GZSL results on CUB and NAB datasets with Wikipedia descriptions as semantic information on the two-split setting. We have used Top-1 \% accuracy and Seen-Unseen AUC (\%) for  ZSL and GZSL, respectively.} 
\resizebox{0.8\textwidth}{!}{
\small\begin{tabular}{|l|cc|cc|cc|cc|}
%\caption{}
\hline
& \multicolumn{4}{c|}{zero-shot Learning} & \multicolumn{4}{c|}{Generalized zero-shot Learning}\\
\cline{2-9}
Methods & \multicolumn{2}{c|}{CUB} & \multicolumn{2}{c|}{NAB} & \multicolumn{2}{c|}{CUB} & \multicolumn{2}{c|}{NAB}  \\
\cline{2-9}
 & Easy & Hard & Easy & Hard & Easy & Hard & Easy & Hard \\
\hline %\cline{2-13}
WAC-Linear~\cite{elhoseiny2013write} & 27.0 & 5.0 & - & - & 23.9 & 4.9 & 23.5 & -   \\
%\hline
WAC-Kernal~\cite{elhoseiny2016write}  & 33.5 & 7.7 & 11.4 & 6.0 & 14.7 & 4.4 & 9.3 & 2.3  \\
%\hline
ESZSL~\cite{romera2015embarrassingly} & 28.5 & 7.4 & 24.3 & 6.3 & 18.5 & 4.5 & 9.2 & 2.9   \\     
%\hline
ZSLNS~\cite{Qiao2016} & 29.1 & 7.3 & 24.5 & 6.8 & 14.7 & 4.4 &9.3 & 2.3    \\
Sync-fast \cite{changpinyo2016synthesized} & 28.0 & 8.6 & 18.4 & 3.8 & 13.1 & 4.0 & 2.7 & 3.5   \\
%Sync-ovo \cite{changpinyo2016synthesized} & 12.5 & 5.9 & - & - & 1.7 & 1.0 & 0.1 & -   \\
%\hline
ZSLPP~\cite{Elhoseiny_2017_CVPR} & 37.2 & 9.7 & 30.3 & 8.1 & 30.4 & 6.1 & 12.6 & 3.5  \\
GAZSL~\cite{Noisy_Text2} & 43.7 & 10.3 & 35.6 & 8.6 & 35.4 & 8.7 & 20.4 & 5.8    \\     
%\hline
LsrGAN [ours] & \textbf{45.2} & \textbf{14.2} & \textbf{36.4} & \textbf{9.04} & \textbf{39.5} & \textbf{12.1} & \textbf{23.2} & \textbf{6.4}  \\
\hline
\end{tabular}
}
%\vspace{-17pt}
\end{table*}

\subsection{ZSL and GZSL Performance}
% zero-shot learning performance 
The results for the ZSL are provided in the left part of Table  2 and 3.  It can be seen that our LsrGAN achieves superior performance in both attribute and Wikipedia based datasets compared to the previous state of the art models, especially with generative models GAZSL, F-GAN, cycle-CLSWGAN, and LisGAN. It is worth noticing that all the mentioned generative models have the same base architecture. e.g., WGAN. Hence, the superiority of our model suggests that our motivation is realistic, and our experiments are effective. In summary, we achieve, 1.5 \%, 3.9 \%, 0.8 \%, 0.44\% improvement on CUB (Easy), CUB (Hard), NAB (Easy) and NAB (Hard) respectively for the Wikipedia based datasets under ZSL. On the other hand, for the attribute-based ZSL, we attain 1.5\% and 0.8\% improvement on CUB and SUN, respectively. ZSL under performs on AWA probably due to the high feature correlation between similar unseen classes having a common neighbor among the seen classes, e.g. Dolphin and Blue whale. The availability of similar unseen classes slightly affects the prediction capability of the classifier for ZSL as the SR-loss tends to cluster them together due to high semantic similarity with common seen classes. 
However, it is worth noticing that the GZSL result for the same dataset is superior. \\
\indent Our primary focus is to elevate the GZSL performance in this work, which is apparent from the right side of Table 2 and 3. Following \cite{good_ugly, Noisy_Text2, F-WGAN}, we report the harmonic mean and AUC score for the attribute and Wikipedia based datasets, respectively. The mentioned metrics help us to showcase the approach's generalizability as the harmonic mean and AUC scores are only high when the performance on seen and unseen classes is high. From the results, we can see that LsrGAN outperforms the previous state of the art for the GZSL. In terms of numbers, we achieve,  4.1\%, 3.4\% 2.8\% and 0.6\% gain on  CUB (Easy), CUB (Hard), NAB (Easy) and NAB (Hard) respectively for the Wikipedia based datasets and 0.7 \%, 1.4\% and 0.7 \% improvement on attribute-based AWA, CUB and SUN respectively. It is worth noticing that the LsrGAN improves the unseen Top-1 performance in the GZSL setting for the attribute-based CUB and SUN by 1.6\% and 1.9\% with the previous state of the art LisGAN \cite{LGAN}. \\
\indent The majority of the conventional approaches, including the generative models overfit the seen classes, which results in lower GZSL performance. Notice that most of the approaches mentioned in Table 2 achieve very high performance on seen classes compared to the unseen classes in the GZSL setting. For example, SYNC \cite{changpinyo2016synthesized} has around 90\% recognition capability on the seen classes, and it drops to only 10\%  (80\% difference) for the unseen classes on AWA dataset. It is also evident from Tables 2 and 3 that the performance on the unseen categories drops drastically when the search space includes both seen and unseen classes in the GZSL setting. For instance, DAP \cite{LNH13} drops from 40 \% to 1.7 \%, GAZSL \cite{Noisy_Text2} drops from 55.8\% to 23.9\% and F-GAN \cite{F-WGAN} drops from 57.3\% to 43.7 \% on attribute-based CUB dataset. This indicates that the previous approaches are easy to overfit the seen classes. Although the generative models have achieved significant progress compared to the previous embedding methods, they still possess the overfitting issue towards seen classes by having substandard GZSL performance. On the contrary, our model incorporates the novel SR-Loss that enables the explicit knowledge transfer from the similar seen classes to the unseen classes. Therefore, the proposed LsrGAN alleviates the overfitting concern and helps us to achieve a state of the art GZSL performance. It is worth noticing that our model  not only outperforms the generative zero-shot models having single GAN but also proves its worth against cycle-GAN \cite{CycleGAN1} in ZSL and GZSL setting. %As explained above, we owe the success of our LsrGAN model to the SR-loss for enabling explicit knowledge transfer from similar seen classes to the unseen classes.  
Lastly, to have fair comparisons, we have taken the performance numbers from \cite{good_ugly, F-WGAN, Noisy_Text2}.

\subsection{Effectiveness of SR-Loss}
\begin{figure}[t]
\centering
\includegraphics[width=0.8\textwidth,height=2.4cm]{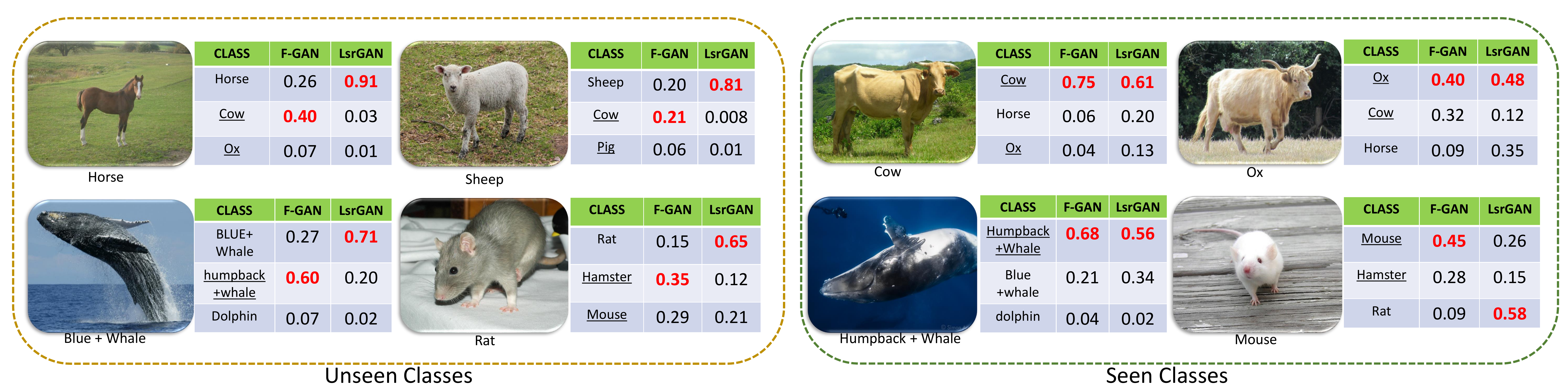}
\caption{Average class confidence score (Average SoftMax Probability) comparison for classifier trained with F-GAN and LsrGAN. Top 3 average guesses are mentioned here. The red marked label showcase the top 1 average guess. The class names with underline represent seen classes.}
%\vspace{-16pt}
\end{figure}
Utilizing the semantic relationship between seen and unseen classes to infer the visual characteristics of an unseen class is at the heart of the proposed SR-Loss. Contrary to other generative approaches, it enables explicit knowledge transfer in the generative model to make it learn from the unseen classes together with seen classes during the training process itself. As a result, our LsrGAN will become more robust towards the unseen classes leading to address the seen class overfitting concern. To demonstrate such ability, we have computed the average class confidence score (Average Softmax probabilities) of the classifier trained with the generated features from the LsrGAN or F-GAN model. Since the confidence scores are computed under the GZSL, the classifier's training set contains real visual features from the seen classes and generated unseen features from LsrGAN or F-GAN. To have a fair comparison, we have used the same F-GAN's Softmax classifier in LsrGAN. Since LsrGAN learns from the seen and unseen classes during the training process itself, the Softmax classifier associated with it is not trained in an offline fashion like in the F-GAN model. \\
\indent Fig. 3 depicts the confidence results on the AWA dataset under the GZSL setting. We have taken mainly four confusing seen and unseen classes for the comparison with classifier's top 3 guesses.  It is evident from the figure that the classifier trained with the F-GAN has lower confidence for the unseen classes, and it mainly showcases very high confidence towards the similar seen classes even if the test image comes from the unseen classes. Also, during the seen class classification, the classifier's confidence values are mainly distributed among the seen classes in F-GAN. For instance, ``mouse'' has its confidence spread between mouse and hamster only. On the other hand, LsrGAN showcases decent confidence values for the seen and unseen, both leading to better GZSL performance.  It is worth noticing that the LsrGAN fails for the ``mouse'' classification. However, the confidence is well spread among all the three categories showing it has considered an unseen class ``rat'' with other seen classes ``mouse" and ``Hamster''. These observations reflect the fact that F-GAN has an overfitting issue towards the seen classes. On the other hand, the balanced performance of LsrGAN manifests that explicit knowledge transfer from SR-loss helps it to overcome the overfitting concern towards the seen classes. To bolster our claim, we have also computed the average class confidence across all the seen and unseen classes for these two models on attribute-based AWA, CUB, and SUN datasets. Table 4 reports average confidence values for seen and unseen classes. Clearly, it shows the superiority of LsrGAN in terms of generalizability compared to F-GAN.

% That weight and epsilon graph, I will mention it here. Will say that how important is the knowledege transfer and mentoin my story !! 

\begin{table*}[t]
\centering
\caption{Comparison of average class confidence score across all seen or unseen classes (SoftMax Probability) between F-GAN and LsrGAN for attribute-based AWA, CUB and SUN} 
\label{tab:gzsl2}
\resizebox{0.45\textwidth}{!}{
\small\begin{tabular}{|l|cc|cc|}
\hline
\multirow{4}{*}{} & \multicolumn{2}{c|}{F-GAN \cite{F-WGAN}} & \multicolumn{2}{c|}{LrsGAN [ours]}\\
\hline
%& \multicolumn{2}{c|}{CUB} & \multicolumn{2}{c|}{NAB} & \multicolumn{2}{c|}{CUB} & \multicolumn{2}{c|}{NAB}  \\
 & Unseen & Seen & Unseen & Seen \\
\hline %\cline{2-13}
AWA &  0.29 & 0.86 & 0.69 & 0.83 \\
CUB &  0.33 & 0.65 & 0.60 & 0.64 \\
SUN &  0.32 & 0.35 & 0.65 & 0.39  \\
\hline
\end{tabular}
}
\end{table*}

\subsection{Model Analysis}
\textbf{Parameter Sensitivity :}
We have tuned our parameters following the conventional grid search approach. We mainly tune the SR-Loss parameters $\epsilon$, $\lambda_{sr}$ and $n_c$. For the fair comparison, we have adopted other parameters $\lambda_{vp}$, $\lambda_{gp}$ from \cite{F-WGAN, Noisy_Text2}, also the $\lambda_{c}$ is considered between $(0,1]$, specifically, $0.01$ for the majority of our experiments. Fig. 4(b)-(d) show the parameter sensitivity for the SR-Loss. Notice that we estimate the $\epsilon$ value from the seen class visual and semantic relations. It can be seen that a lower and higher value of $\epsilon$ affects the performance. On the other hand, $\lambda_{sr}$ and $n_c$ maintain consistent performance after reaching a certain threshold value.

\begin{figure*}[t!h]
\begin{center}
\subfigure[{Ablation study}]{
\includegraphics[width=0.25\textwidth, height=0.17\linewidth]{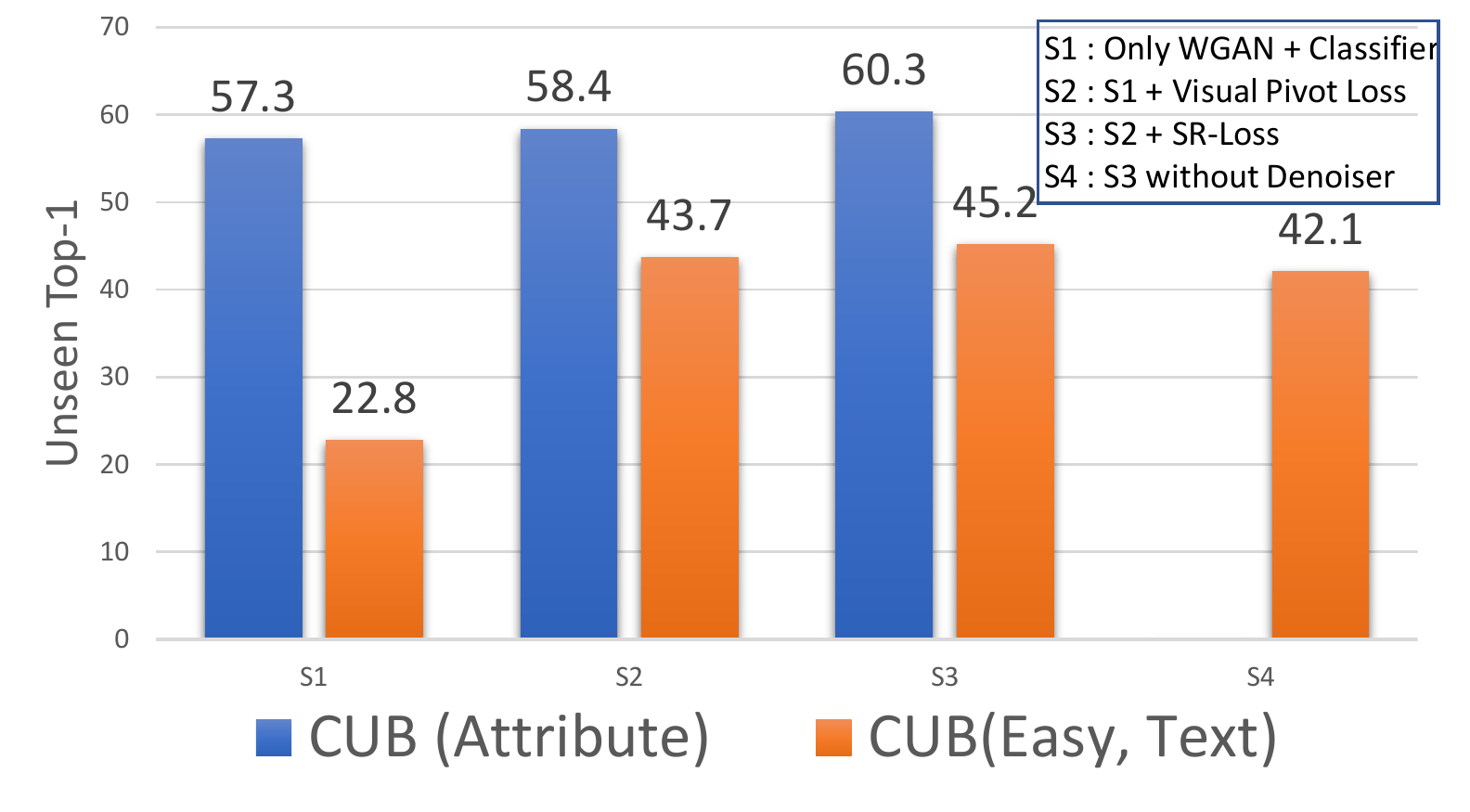}
}
\subfigure[{$\epsilon$ (CUB E)}]{
\includegraphics[trim = 2mm 1mm 10mm 9mm, clip, width=0.20\textwidth]{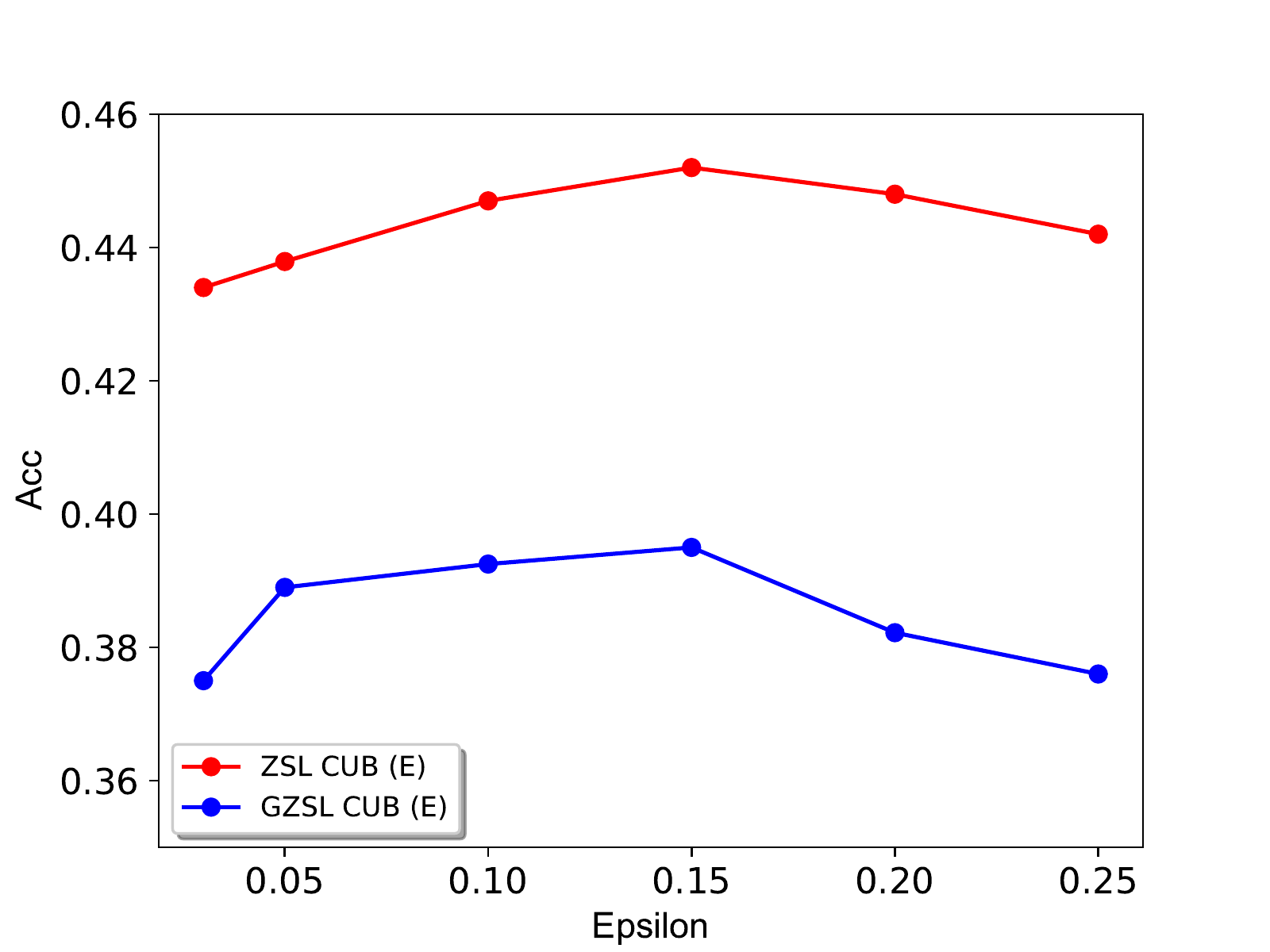}
}
\subfigure[{$\lambda_{sr}$ (AWA)}]{
\includegraphics[trim = 2mm 1mm 10mm 9mm, clip,width=0.20\textwidth, height=0.148\linewidth]{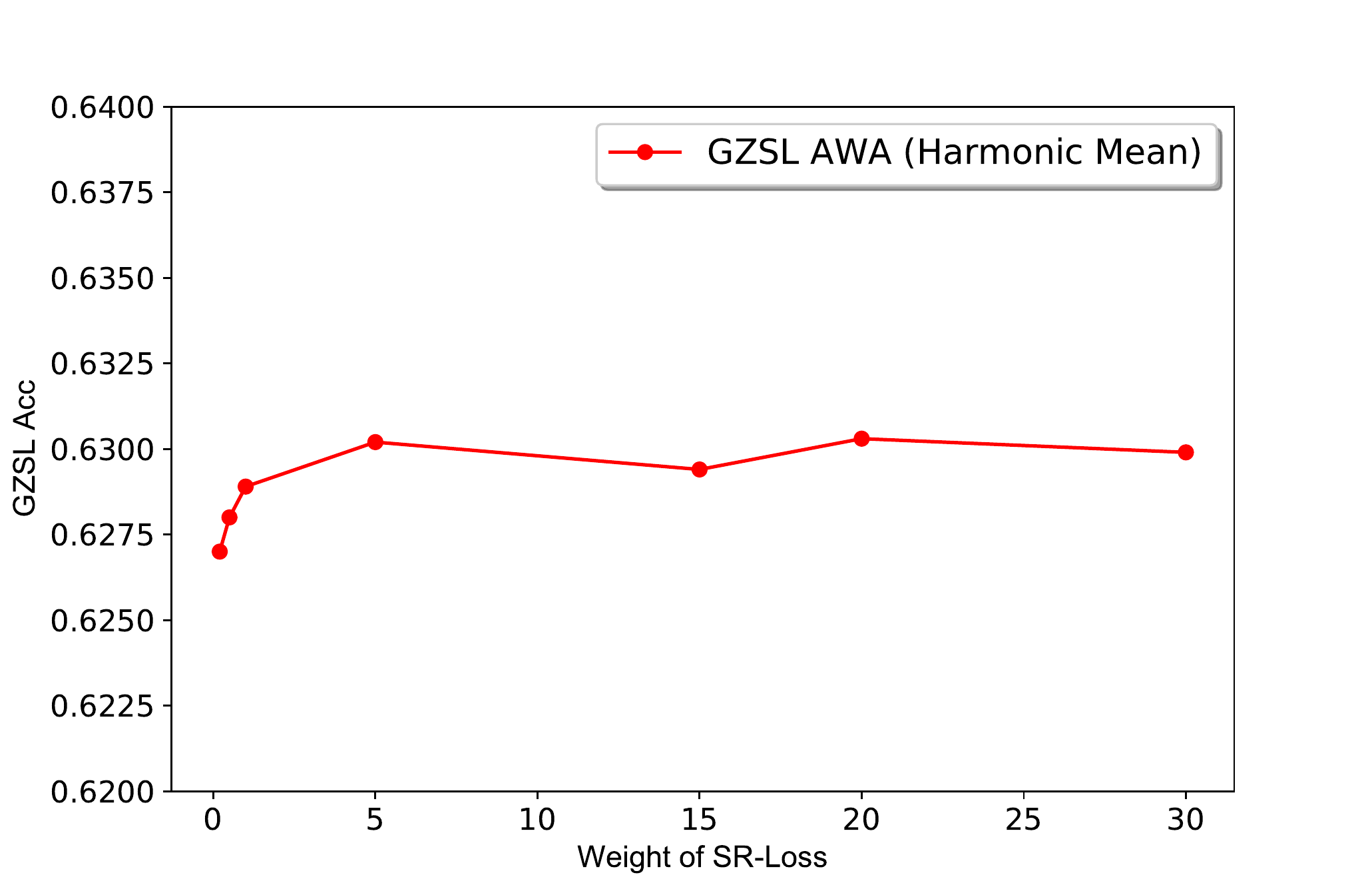}
}
\subfigure[{$n_c$ (CUB E)}]{
\includegraphics[trim = 2mm 1mm 10mm 9mm, clip, width=0.20\textwidth]{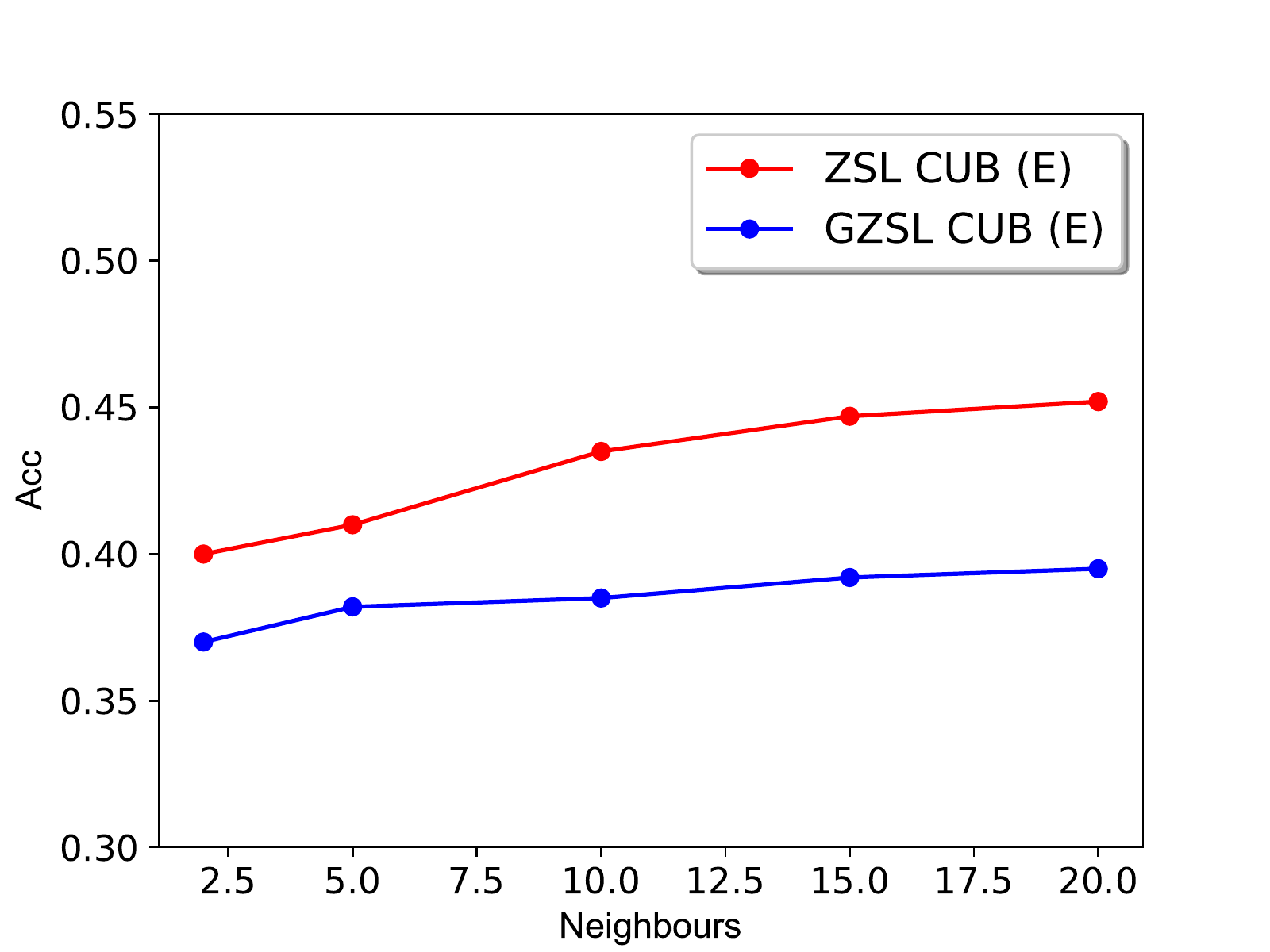}
}
\end{center}
%\vspace{-5pt}
\caption{Ablation study under ZSL(a), Parameter sensitivity  for the SR-loss (b-d).}
\label{fig:para}
\end{figure*}

\begin{figure*}[t!h]
\begin{center}
\subfigure[{ZSL  Text}]{
\includegraphics[width=0.30\linewidth, height=0.20 \linewidth]{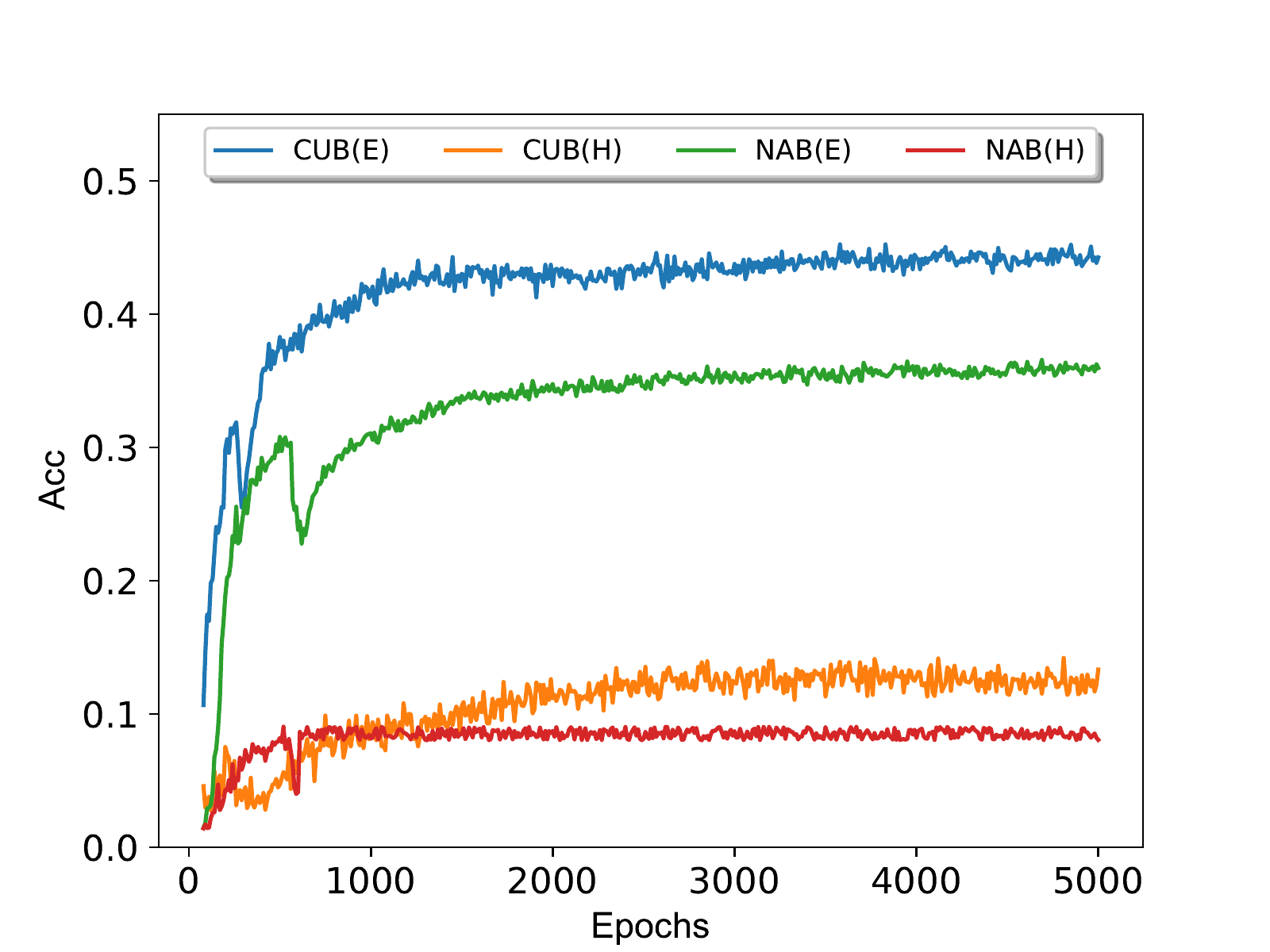}
}
\subfigure[{GZSL Text}]{
\includegraphics[width=0.30\linewidth, height=0.20 \linewidth]{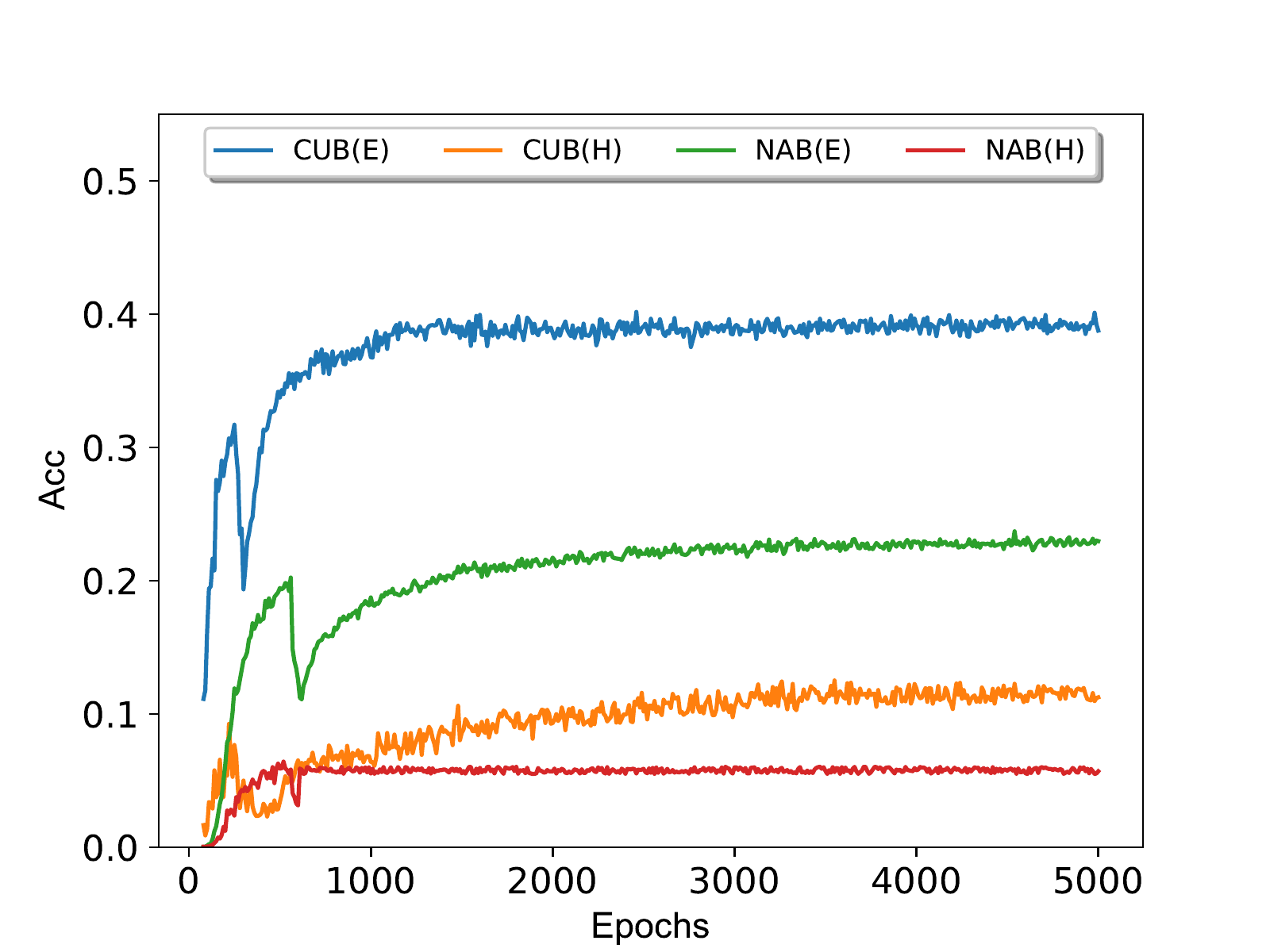}
}
\subfigure[{ZSL-GZSL Attribute}]{
\includegraphics[width=0.30\linewidth, height=0.20 \linewidth]{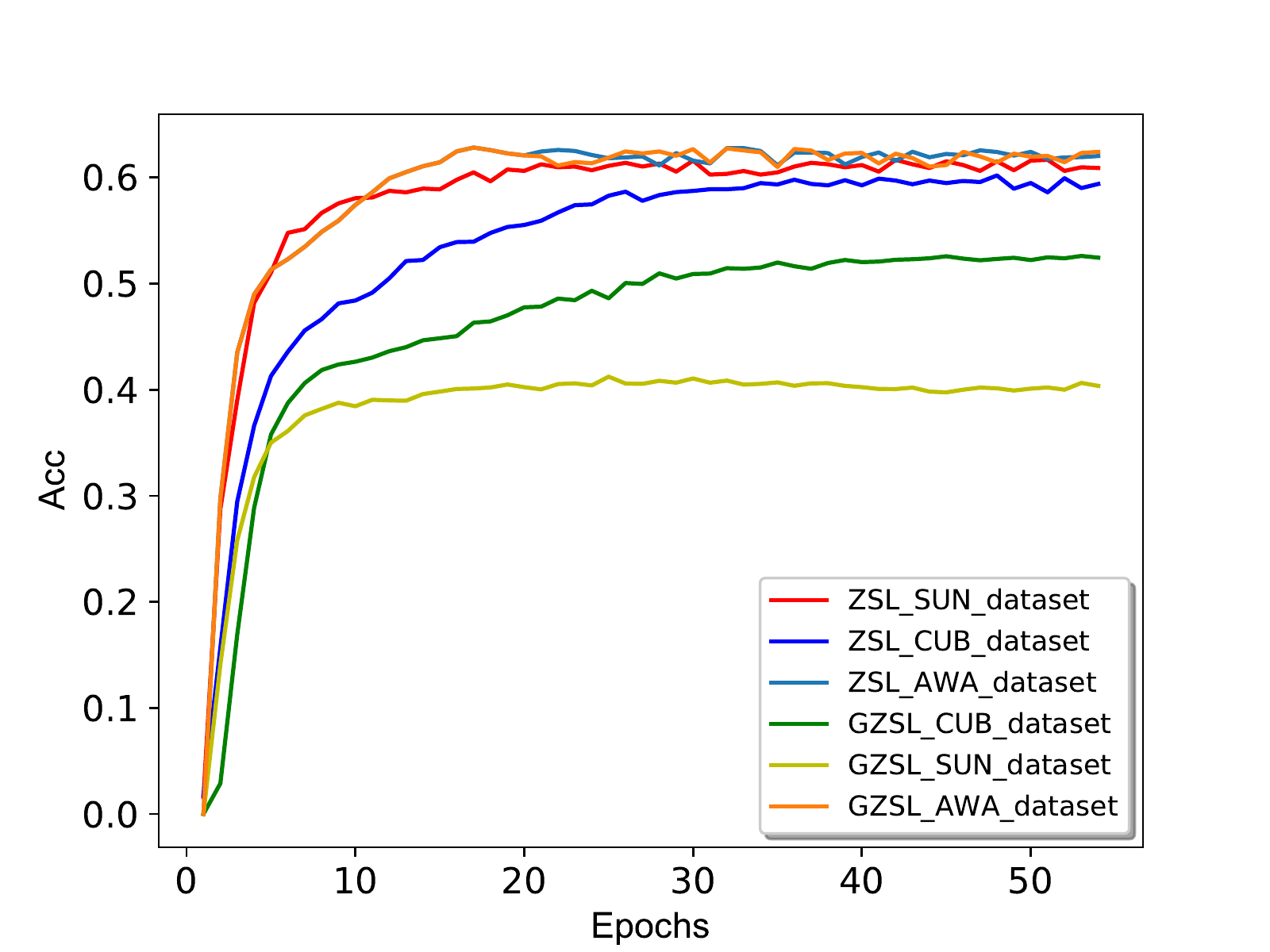}
}
\end{center}
%\vspace{-5pt}
\caption{Training stability (a-c) across all datasets under ZSL and GZSL}
\label{fig:para2}
\end{figure*}

\textbf{Training Stability :}
Since GANs are notoriously hard to train, and our proposed LsrGAN model not only uses GAN but also optimizes the similarity constraints from the SR-loss. Therefore, we also report the training stability for ZSL and GZSL for attribute and Wikipedia based datasets. Specifically, for the ZSL, we report the unseen Top-1 accuracy and Epoch behavior in Fig. 5(a) and Fig. 5(c). We have used harmonic mean of the seen and unseen Top-1 accuracy and Epoch to showcase the GZSL stability, mentioned in Fig. 5(b) and Fig. 5(c). Mainly we see the stable performance across all the datasets. 

\textbf{Ablation study :}
We have reported the Ablation study of our model in Fig. 4 (a) under ZSL for both attribute and Wikipedia based CUB. Primarily, we have used CUB (Easy) split for the Wikipedia based dataset. The $S1$ - WGAN with a classifier is considered as a baseline model. $S2$ reflects the effect of the visual pivot regularizer in our model. Finally, $S3$ showcase the performance of a complete LsrGAN with SR-loss. To highlight the effect of the denoiser used to process the Wikipedia based features, we have also reported the LsrGAN without denoiser in $S4$. In summary, Fig. 4 (a) showcases the importance of each component used in our model.

\section{Conclusions}

In this paper, we have proposed a novel generative zero-shot model named LsrGAN that Leverages the Semantic Relationship between seen and unseen classes to address the seen class overfitting concern in GZSL. Mainly, LsrGAN employs a novel Semantic Regularized Loss (SR-Loss) to perform explicit knowledge transfer from seen classes to unseen ones. The SR-Loss explores the semantic relationships between seen and unseen classes to guide the LsrGAN to generate visual features that mirror the same relationship. Extensive experiments on seven benchmarks, including attribute and Wikipedia description based datasets, verify that our LsrGAN effectively addresses the overfitting concern and demonstrates a superior ZSL and GZSL performance.  

\section{Appendix}

\subsection{Time complexity of the SR-loss}

\begin{enumerate}
   \item \textbf{For the semantic similarity:}
   \begin{itemize}
     \item Cosine Similarity matrix ($n\times n$) : $\mathcal{O}(n^2 L)$, $L$ length of semantic feature, $n$ total classes
    \item To get the $K$ similar classes : $\mathcal{O}(n^2 \log{} K)$, $\mathcal{O}(n \log{} K)$ for each  class, heap sort.
    \item Overall cost :  $\mathcal{O}(n^2 L)$ + $\mathcal{O}(n^2 \log{} K)$ 
   \end{itemize}
   
   \item \textbf{For the visual similarity:} 
   \begin{itemize}
       \item Cosine Similarity matrix ($B\times K$) : $\mathcal{O}(B K V)$, $V$ length of visual feature, $B$ batch size classes, $K$ neighbour classes. 
        \item Overall cost :   $\mathcal{O}(B K V E)$, $E$ total number of epochs.
   \end{itemize} 
    
\end{enumerate}

\begin{itemize}
    \item Time Complexity SR-loss (Clean Attributes): $\mathcal{O}(B K V E)$ + $\mathcal{O}(n^2 L)$ + $\mathcal{O}(n^2 \log{} K)$ 
    \item Time Complexity SR-loss (Noisy Text): $\mathcal{O}(B K V E)$ + $\mathcal{O}(n^2 L E)$ + $\mathcal{O}(n^2 \log{} K E)$
\end{itemize}

\noindent Clearly, the overall time complexity is linear in terms of $E$, $K$ , $B$, $V$, and $L$ and degree 2 polynomial with $log$ in terms of the total classes. It is worth noticing that the complexity is not exponential and the running time cost is manageable. 

\pagebreak 

\subsection{Training Algorithm}
Below, we illustrate our training procedure for the LsrGAN model. We train the Generator ($G_\theta$) and Discriminator ($D_\theta$) alternately with the Adam optimizer. Notice that the training of LsrGAN contains two phases, one for the seen classes and another for the unseen classes.

\begin{algorithm}[]
	\begin{algorithmic}[1]
% 		\STATE{\bfseries Input:} the maximal loops $N_{step}$, the batch size $m$, the iteration number of discriminator in a loop $n_{d}$, the balancing parameter $\lambda_p$, Adam hyperparameters $\alpha_1$, $\beta_1$, $\beta_2$. 
        \STATE{\bfseries Input:} number of epochs $N_{E}$, the batch size $m$, discriminator iterations $N_{d}=5$ for seen classes, loss hyper parameters $\lambda_c$, $\lambda_{vp}$ and $\lambda_{sr}$, $N_{c}=$ $1$ or $2$ discriminator iterations for unseen classes, and  Adam parameters $\beta_1=0.5$ and $\beta_2=0.9$. 

        \FOR{iter $= 1,..., N_{E}$} 
        \STATE \textit{// Seen Class Training} 
        % \STATE \textcolor{blue}{Sample random text minibatches $t_a, t_b$, noise $z^h$}
        % \STATE \textcolor{blue}{Construct $t^h$ using Eq.6 with different $\alpha$ for each row in the minibatch}
        % \STATE \textcolor{blue}{$\tilde{x}^h \gets G(t^h, z^h)$}
		\FOR{$i = 1$, ..., $N_{d}$}
		\STATE Minibatch sampling from $\cT^s$ with matching images from $\cX^s$ and noise $\cZ$ 
        %\STATE Sample a minibatch of images $x$,  matching texts $t$, random noise $z$
		\STATE $\tilde{\bmx}   \leftarrow G_{\theta_g}(\bmt^s, \cZ)$
		\STATE Discriminator and classifier loss computation $\cL_d$ and $\cL_c$ using Eq. 2 and 3 
        \STATE $\theta_d \leftarrow \text{Adam}(\nabla_{\theta_d^r} \cL_d, \bigtriangledown_{\theta_d^c} \cL_c, \theta_d,  \lambda_c, \beta_1, \beta_2)$
        \ENDFOR

        \STATE Minibatch sampling from $\cT^s$ and noise $\cZ$
        \STATE Generator loss computation $L_G$ using Eq. 8
        \STATE $\tilde{\bmx}   \leftarrow G_{\theta_g}(\bmt^s, \cZ)$
		\STATE\begin{varwidth}[t]{\linewidth}  $\theta_g \leftarrow \text{Adam}(\nabla_{\theta_g} \cL_d, \bigtriangledown_{\theta_g} \cL_{vp}, \bigtriangledown_{\theta_g} \cL_{c}, \bigtriangledown_{\theta_g} \cL_{sr}^s,	\theta_g, \lambda_c, \lambda_{vp}, \lambda_{sr}, \beta_1, \beta_2)$ 	
		\end{varwidth}
% 		\STATE\begin{varwidth}[t]{\linewidth}  $\theta_E \gets \text{Adam}(\bigtriangledown_{\theta_E}  L_G ,	\theta, \alpha_1, \beta_1, \beta_2)$
% 		\end{varwidth}
        \vspace{8pt}
        \STATE \textit{// Unseen Class Training} 
        % \STATE \textcolor{blue}{Sample random text minibatches $t_a, t_b$, noise $z^h$}
        % \STATE \textcolor{blue}{Construct $t^h$ using Eq.6 with different $\alpha$ for each row in the minibatch}
        % \STATE \textcolor{blue}{$\tilde{x}^h \gets G(t^h, z^h)$}
		\FOR{$i = 1$, ..., $N_{c}$}
		\STATE Minibatch sampling from $\cT^u$ and noise $\cZ$ 
        %\STATE Sample a minibatch of images $x$,  matching texts $t$, random noise $z$
		\STATE $\tilde{\bmx}   \gets G_{\theta_g}(\bmt^u, \cZ)$
		\STATE Classifier loss computation $\cL_c$ using Eq. 3 
        \STATE $\theta_d^c \leftarrow \text{Adam}(\nabla_{\theta_d^c} \cL_c, \theta_d^c, \lambda_c, \beta_1, \beta_2)$
        \ENDFOR
        \STATE Minibatch sampling from $\cT^u$ and noise $\cZ$
        \STATE Generator loss computation $L_G$ using Eq. 8 
        \STATE $\tilde{\bmx}   \leftarrow G_{\theta_g}(\bmt^u, \cZ)$
		\STATE\begin{varwidth}[t]{\linewidth}  $\theta_g \leftarrow \text{Adam}(\nabla_{\theta_g} \cL_{c}, \bigtriangledown_{\theta_g} \cL_{sr}^u,	\theta_g, \lambda_c, \lambda_{sr}, \beta_1, \beta_2)$ 	
		\end{varwidth}
		\ENDFOR
	\end{algorithmic}
	\caption{Training procedure for the LsrGAN}
	\label{alg_can_label}
% 	\vspace{-3em}
\end{algorithm}

\clearpage
\bibliographystyle{splncs04}
\bibliography{egbib}
\end{document}